\documentclass[10pt,twocolumn,letterpaper]{article}

\usepackage{cvpr}              

\usepackage{graphicx}
\usepackage{amsmath}
\usepackage{amssymb}
\usepackage{booktabs}
\usepackage{multirow}
\usepackage[accsupp]{axessibility} 

\usepackage[pagebackref,breaklinks,colorlinks]{hyperref}

\usepackage[capitalize]{cleveref}
\crefname{section}{Sec.}{Secs.}
\Crefname{section}{Section}{Sections}
\Crefname{table}{Table}{Tables}
\crefname{table}{Tab.}{Tabs.}

\def\cvprPaperID{21} 
\def\confName{CVPR workshop}
\def\confYear{2024}

\begin{document}


\title{LaPA: Latent Prompt Assist Model For Medical Visual Question Answering}

\author{Tiancheng Gu\\
University of Sydney\\
Sydney, NSW, Australia\\
{\tt\small tigu8498@uni.sydney.edu.au}
\and
Kaicheng Yang\\
DeepGlint\\
Beijing, China\\
{\tt\small kaichengyang@deepglint.com}
\and
Dongnan Liu\\
University of Sydney\\
Sydney, NSW, Australia\\
{\tt\small dongnan.liu@sydney.edu.au}
\and
Weidong Cai\\
University of Sydney\\
Sydney, NSW, Australia\\
{\tt\small tom.cai@sydney.edu.au}
}
\maketitle

\begin{abstract}
   Medical visual question answering~(Med-VQA) aims to automate the prediction of correct answers for medical images and questions, thereby assisting physicians in reducing repetitive tasks and alleviating their workload. Existing approaches primarily focus on pre-training models using additional and comprehensive datasets, followed by fine-tuning to enhance performance in downstream tasks. However, there is also significant value in exploring existing models to extract clinically relevant information. In this paper, we propose the \textbf{La}tent \textbf{P}rompt \textbf{A}ssist model~(LaPA) for medical visual question answering. Firstly,  we design a latent prompt generation module to generate the latent prompt with the constraint of the target answer. Subsequently, we propose a multi-modal fusion block with latent prompt fusion module that utilizes the latent prompt to extract clinical-relevant information from uni-modal and multi-modal features. Additionally, we introduce a prior knowledge fusion module to integrate the relationship between diseases and organs with the clinical-relevant information. Finally, we combine the final integrated information with image-language cross-modal information to predict the final answers. Experimental results on three publicly available Med-VQA datasets demonstrate that LaPA outperforms the state-of-the-art model ARL, achieving improvements of 1.83\%, 0.63\%, and 1.80\% on VQA-RAD, SLAKE, and VQA-2019, respectively. The code is publicly available at $\href{https://github.com/GaryGuTC/LaPA_model}{https://github.com/GaryGuTC/LaPA\_model}$.
\end{abstract}

%
\section{Introduction}
\label{sec: Introduction}
Medical visual question answering~(Med-VQA) plays a critical role in disease detection and diagnosis. In clinical practice, the review of numerous medical images and their corresponding questions by physicians is both costly and error-prone~\cite{lin2023medical}. To address this challenge, there has been a growing interest in the development of automatic Med-VQA techniques~\cite{MEVF_model, m3ae, q2atransformer, MMBert_model, vqamix, zhang2022type}. While deep learning models have achieved remarkable success in predicting accurate answers in standard visual-question answering~(VQA) tasks by given images and questions~\cite{BAN_model, SAN_model}, Med-VQA poses unique challenges~\cite{m3ae}. The size of Med-VQA datasets is relatively small, and medical images are complex and challenging due to the small region of interest related to the disease that physicians need to focus on~\cite{COMG_model, rgrg_model}. Consequently, extracting clinically relevant information from medical images becomes a difficult task for the model~\cite{medicat}.


Numerous Med-VQA methods~\cite{m3ae, q2atransformer, MMBert_model, zhang2022type} have been proposed to address the aforementioned challenges and have demonstrated impressive performance. For instance, methods such as MEVF model~\cite{MEVF_model}, MMQ model~\cite{MMQ_model}, and CPCR~\cite{CPCR} have proposed pretraining the model using external complementary datasets to enhance the model's analytical capabilities, followed by fine-tuning for downstream tasks. Similarly, M2I2 model~\cite{M2I2_model} and m3ae model~\cite{m3ae} have utilized self-supervised learning to enable the model to autonomously learn clinical features from both image and language modalities. Notably, despite their remarkable achievements, none of these approaches consider the latent prompt. However, the latent prompt is a crucial aspect that warrants research attention due to its enhanced flexibility in information extraction, as evidenced by its widespread utilization in the field of natural language processing~\cite{ppt_model, Prompt_for_extraction, lpt_ref}.

\begin{figure*}[t]
\begin{center}
\includegraphics[width=1\linewidth]{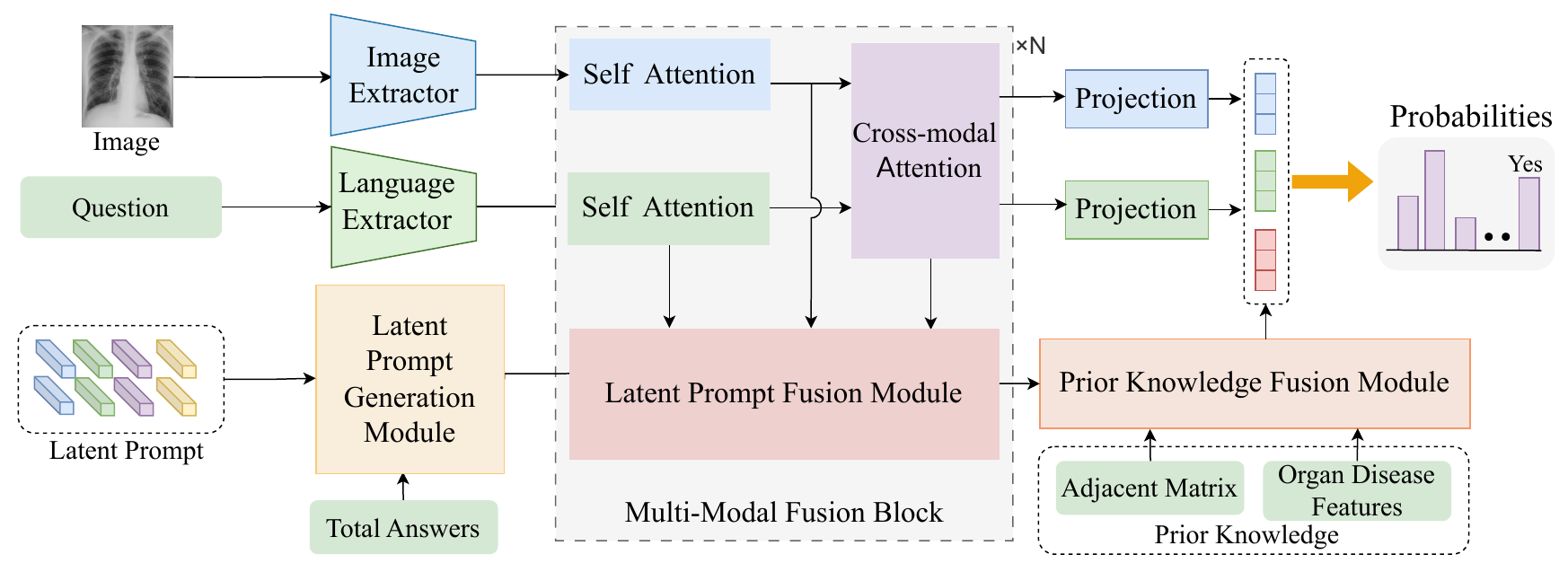}
\end{center}
\caption{The overall structure of our proposed LaPA model. The input feature is denoted by a block with rounded corners, while the square-angled structure represents a module. The language and image pipelines are represented by green and blue modules, respectively. The final tokens in blue, green, and red correspond to the cross-modal image, language, and integrated information, respectively. For optimal viewing, it is recommended to zoom in for detailed examination.}
\label{Fig: The structure of the LaPA}
\end{figure*}



This study presents the LaPA~(Latent Prompt Assist) model for medical visual question answering~(Med-VQA), as illustrated in Fig.~\ref{Fig: The structure of the LaPA}. The LaPA model incorporates the latent prompt to filter different modal information and extract clinic-relevant information, aiding in the prediction of the final answer. Firstly, we introduce the latent prompt generation module, which generates the latent prompt. The latent prompt interacts with the total answer tokens and is constrained by the target answer tokens to focus on the relevant tokens associated with the target answer. Subsequently, the latent prompt is fed into the multi-modal fusion block to fuse with uni- and multi-modal information, enabling the filtering of different modal information and extraction of clinic-relevant details. Additionally, the latent prompt interacts with the prior knowledge derived from the relationship between organs and diseases, obtained from a knowledge graph~\cite{slake}, resulting in the generation of the final interacted information to further assist in the prediction of the final answer. Lastly, the latent prompt combines with the image-language cross-modal information to produce the final answer.

The main contributions of our work can be summarized as follows:
\begin{itemize}
    \item We propose the latent prompt generation model that generates a latent prompt and utilize a multi-modal fusion block to filter different modal information and extract clinic-relevant information.
    \item We leverage prior knowledge regarding the relationship between organs and diseases by employing a graph neural network to interact with the latent prompt, ultimately assisting in answer prediction.
    \item Our proposed LaPA model demonstrates its effectiveness by achieving exceptional performance on VQA-RAD~\cite{rad-vqa}, SLAKE~\cite{slake}, and VQA-2019~\cite{ImageCLEFVQA-Med2019} datasets.
\end{itemize}

\section{Related Works}
\label{sec: related works}

\paragraph*{Prompt Learning.}

Prompt learning is a research focus aimed at leveraging prompts to enhance various aspects of a model's performance, such as efficiency, flexibility, and knowledge transfer~\cite{ppt_model, conditional_prompt_learning, zhou2022learning}. Recent studies~\cite{Prompt_for_extraction, prompt_extract} have explored the utilization of prompts to extract relevant information from pre-trained models for downstream tasks, yielding promising results. Notably, the ChatExtract method proposed by \cite{prompt_extract} employs engineered prompts to aid in sentence differentiation and data extraction, thereby improving answer accuracy. In contrast, \cite{lpt_ref} focuses on using latent prompts, encompassing controlled and uncontrolled signals, to extract valuable and highly relevant information, thereby enhancing text summarization quality. Building upon these studies, we introduce the concept of latent prompts to the domain of Med-VQA.

\begin{figure*}[t]
\begin{center}
\includegraphics[width=1\linewidth]{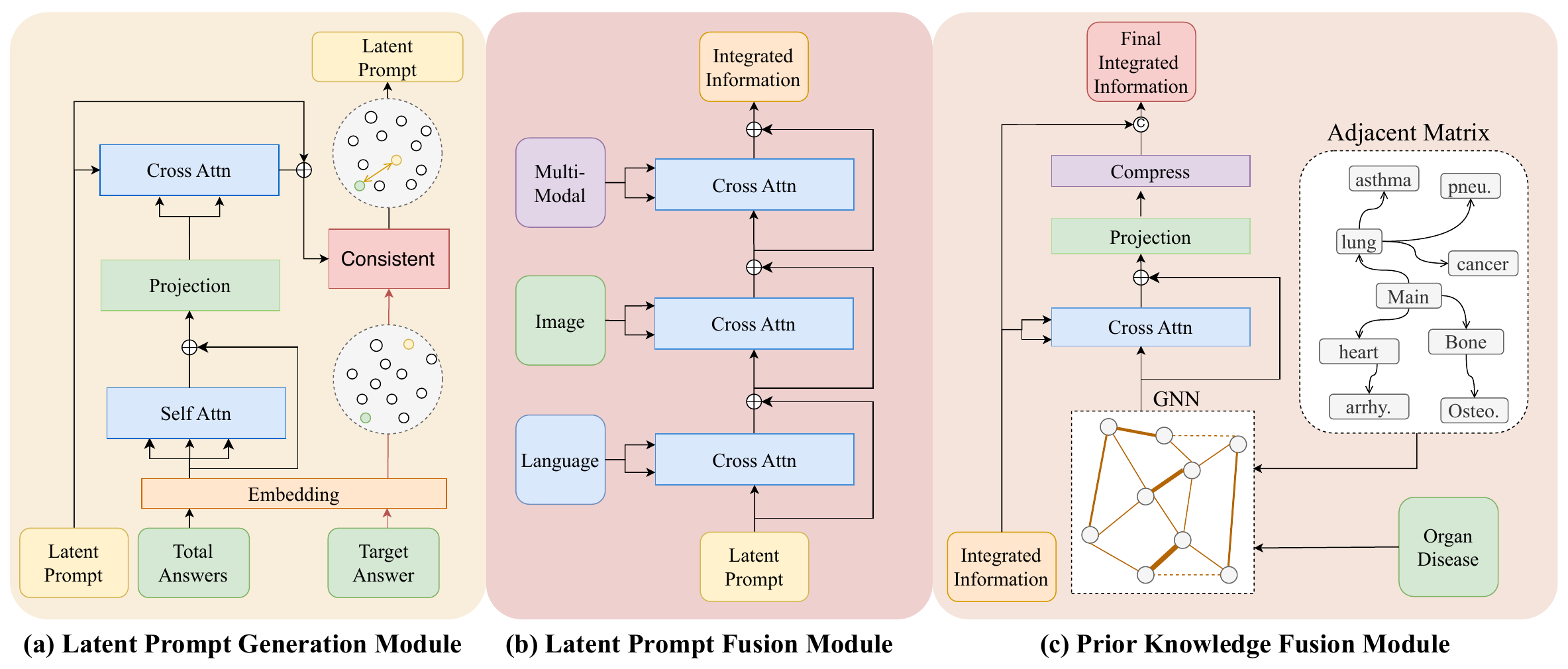}
\end{center}
 \caption{The structure of the main modules in LaPA is illustrated as follows: (a), (b), and (c) represent the latent prompt generation module~(Sec.~\ref{subsec: Latent Prompt Generation Module}), the latent prompt fusion module~(Sec.~\ref{subsec: Multi-modal Feature Fusion Module}), and the prior knowledge fusion module~(Sec.~\ref{subsec: Knowledge-graph-based Latent Prompt Analysis Module}), respectively. For optimal visualization, it is recommended to zoom in for detailed examination.}
\label{Fig: The structure of LaPA's module}
\end{figure*}

\paragraph*{Medical Visual Question Answering.} 

The field of automatic prediction of answers for medical visual questions based on medical images has been extensively studied, yielding numerous notable works~\cite{MEVF_model, q2atransformer, MMBert_model, zhang2022type}. Notably, some approaches have been proposed to train models based on external knowledge, such as MEVF model~\cite{MEVF_model} and MMQ model~\cite{MMQ_model}. These methods initialize the weights of specific modules (e.g., visual encoder or decoders) using pre-trained large language models~(LLMs) and subsequently fine-tune the overall frameworks for downstream Med-VQA tasks. Q2ATransformer~\cite{q2atransformer} introduces a novel approach that combines the advantages of both classification and generation techniques, achieving a unified treatment for closed-end and open-end questions. By employing learnable candidate answer embeddings, Q2ATransformer queries the presence of each answer class for a given image-question pair. Additionally, MeDVInt~\cite{MedVInT} and LLaVA-Med~\cite{llava-med} are generative models for Med-VQA understanding that align visual information from a pre-trained vision encoder with a large language model (LLM) or large vision language model such as ChatGPT and LLaVA. In contrast to these existing works, our proposed approach utilizes latent prompts to filter uni- and multi-modal information and extract clinic-relevant information, thereby enhancing the final answer prediction process.

\section{LaPA Model}
\label{sec: DI-VQA}


The architectural overview of our proposed LaPA~(Latent Prompt Assist) model for medical visual question answering is presented in Fig.~\ref{Fig: The structure of the LaPA}. The model comprises three key components: the latent prompt generation module~(Sec.~\ref{subsec: Latent Prompt Generation Module}), the multi-modal fusion block~(Sec.~\ref{subsec: Multi-modal Feature Fusion Module}), and the prior knowledge fusion module~(Sec.~\ref{subsec: Knowledge-graph-based Latent Prompt Analysis Module}). Further insights into the training process can be found in Sec.~\ref{subsec: training details}.


\subsection{Latent Prompt Generation Module}
\label{subsec: Latent Prompt Generation Module}


We first propose a latent prompt generation module~(Fig.~\ref{Fig: The structure of LaPA's module}~(a)) to generate the learnable latent prompt, which is initialized using the normal distribution. To improve training efficiency and performance, we interact the generated latent prompt with total answer tokens. Under the constraint of answer tokens, the latent prompt can focus on the tokens associated with the answer. To this end, we treat all the answer tokens in the downstream datasets as prior knowledge, embedding them as features $\rm X_{TA}$ using RoBERTa~\cite{roberta_model}. Subsequently, the total answer tokens undergo self-attention followed by a projection layer to obtain the total token features $\rm F_{TA}$ as follows:
\begin{equation}
   \rm F_{TA} = Proj(SA(X_{TA})),
\end{equation}
where $\rm SA(\cdot)$ and $\rm Proj(\cdot)$ represent self-attention mechanism and projection layer respectively. After that, we employ cross-attention to integrate the total answer tokens with the latent prompt:
\begin{equation}
   \rm \hat{X}_{LP} = CA(X_{LP}, F_{TA}, F_{TA}),
\end{equation}
where $\rm CA(\cdot)$ represents the cross-attention mechanism~\cite{transformer_model} with the query, key and value as input. To focus on answers-related tokens, we introduce a consistent loss~$\rm \mathcal{L}_{CS}$ to constrain the latent prompt with the target answer, thereby bringing it closer to the target answer in the semantic space. The process is defined as:
\begin{equation}
    \rm \mathcal{L}_{CS}(\hat{X}_{LP}, X_{A}) = 1- \frac{\hat{X}_{LP}^\top X_{A}}{||\hat{X}_{LP}||~||X_{A}||},
\end{equation}
where $\rm X_{A}$ is the token embeddings of the target answer.

\subsection{Multi-modal Fusion Block}
\label{subsec: Multi-modal Feature Fusion Module}

To make the latent prompt fully extract clinic-relevant information from uni-modal and multi-modal information, we introduce a multi-modal feature fusion block. As shown in Fig.~\ref{Fig: The structure of the LaPA}, the image features and language features are extracted by the Swin Transformer~\cite{swin_transformer} and the RoBERTa~\cite{roberta_model}, and the uni-modal features $\rm F_{I}$ and $\rm F_{L}$ can be obtained through self-attention as follows: 
\begin{equation}
    \rm F_{I} = SA({E_{I}(X_{I})}),
\end{equation}
\begin{equation}
    \rm F_{L} = SA({E_{L}(X_{L})}).
\end{equation}
After that, the image and language features are fused through the cross-attention to get the multi-modal features $\rm F_{MM}$:
\begin{equation}
    \rm F_{MM} = [Proj(CA(F_{I}, F_{L}, F_{L})); Proj(CA(F_{L}, F_{I}, F_{I}))],
\end{equation}
where $\rm Proj(\cdot)$ represents the projection layer. 

After getting the uni-modal and multi-modal features $\rm F_{I}$, $\rm F_{L}$, and $\rm F_{MM}$, we design the latent prompt fission module~(Fig.~\ref{Fig: The structure of LaPA's module}~(b)) to make the latent prompt to integrate clinic-relevant information through cross-attention:


\begin{equation}
    \rm {X}_{II} = CA(\hat{X}_{LP}, F_{I}, F_{I}),
\end{equation}
\begin{equation}
    \rm \check{{X}}_{II} = CA({X}_{II}, F_{L}, F_{L}),
\end{equation}
\begin{equation}
   \rm \Tilde{{X}}_{II} = CA(\check{{X}}_{II}, F_{MM}, F_{MM}),
\end{equation}
where $\rm CA(\cdot)$ is the cross-attention mechanism. The $\rm {X}_{II}$ represents the integrated information obtained by combining latent prompts with image features. Similarly, the $\rm \check{{X}}_{II}$ and the $\rm \Tilde{{X}}_{II}$ denote the integrated information resulting from the fusion of language features and multi-modal features, respectively. The fusion process follows a sequential order, where language features are integrated first, followed by images, and finally multi-modal features. We have conducted experiments to explore various approaches for information fusion and extraction, and the current form yields the most optimal results.

In the multi-modal fusion module, the latent prompt is utilized to integrate with language features to extract clinically relevant information within the textual semantic space. Subsequently, it is combined with image features to extract clinically relevant information within the image semantic space. Finally, the integrated information undergoes fusion with the combined language-image cross-modal features to filter out diverse modal information and consolidate the uni-modal features of both language and image, along with their multi-modal combination features, resulting in the generation of the final clinically relevant information.

\subsection{Prior Knowledge Fusion Module}
\label{subsec: Knowledge-graph-based Latent Prompt Analysis Module}
Following the previous works~\cite{GAT_model, graph_neural_network}, we incorporate a prior knowledge graph~\cite{slake} that captures the relationships between organs and diseases to enhance the accuracy of answer prediction in Med-VQA. We employ a graph neural network~($\rm GNN(\cdot)$) to analyze the organ-disease relationships and improve the performance of answer prediction. Additionally, we propose a prior knowledge fusion module that integrates the prior knowledge with the integrated information to facilitate the final answer prediction.

As depicted in Fig.~\ref{Fig: The structure of LaPA's module}~(c), the adjacent matrix $\rm X_{adj}$ is derived from the aforementioned prior knowledge~\cite{slake}, representing the relationship between organs and diseases using binary values (0 and 1). The organ-disease feature $\rm F_{OD}$ is tokenized and embedded using RoBERTa~\cite{roberta_model}. Subsequently, it is fed into the GNN module to extract valuable information regarding the organ-disease relationships denoted as $\rm F_{G}$, which can be summarized as follows:

\begin{equation}
    \rm F_{G} = GNN(F_{OD}, X_{adj}).
\end{equation}
Then, the extracted information is combined with the previous integrated information~$\rm \widetilde{x}_{LP}$ to get the final integrated information~($\rm \hat{X}_{II}$), and the process is indicated below:
\begin{equation}
    \rm \hat{X}_{II} = [\Tilde{{X}}_{II};Proj(CA(F_{G}, X_{LP}, X_{LP}))],
\end{equation}
where $\rm CA(\cdot)$ is the cross attention mechanism  and $\rm Proj(\cdot)$ is the projection layer. Finally, the interacted relationship-based features will concatenate~($[;]$) with latent prompt as the final integrated information to assist the final answer predicted for Med-VQA.




\subsection{Training Details}
\label{subsec: training details}
After the processes mentioned above, we add the cross-modal information~$\rm F_{FI}$ and~$\rm F_{FL}$ of the cross-modal attention in the last multi-modal fusion block with the final integrated information~$\rm \hat{X}_{II}$ to predict the answer:
\begin{equation}
    \rm X_{F} = \alpha \hat{X}_{II} + \theta F_{FI} + \beta F_{FL},
\end{equation}
where $\alpha$, $\theta$, and $\beta$ are weight to balance different types of information. This final total loss~($\rm \mathcal{L}_{T}$) is shown below:
\begin{equation}
    \rm \mathcal{L}_{T} = \mathcal{L}_{BCE}(X_{F}, F_{T}) + \eta \mathcal{L}_{CS},
\end{equation}
where the $\rm \mathcal{L}_{BCE}$ is the binary cross-entropy loss~\cite{BCE_loss} and $\rm \mathcal{L}_{CS}$ is the consistent loss used to minimize the semantic distance between the latent prompt and the target answer. $\rm \eta$ is a loss weight to adjust the influence of different losses.



\section{Experiments and Results}
\label{sec: experiments and results}

\begin{table*}[ht]
    \centering
    \begin{tabular}{ c l | c c c | c c c |  c}
        \hline \hline
        \multirow{2}{*}{Method} &\multirow{2}{*}{Venue} 
        &\multicolumn{3}{c}{VQA-RAD}\vline
        & \multicolumn{3}{c}{SLAKE}\vline
        & \multicolumn{1}{c}{VQA-2019} \\
        & & Open & Closed & Overall & Open & Closed & Overall & Overall \\
        \midrule
        BAN~\cite{BAN_model} & NeurIPS$_{18}$ & 37.40 & 72.10 & 58.30 & 74.60 & 79.10 & 76.30 & - \\
        CPRD-BAN~\cite{CPRD_model} & MICCAI$_{21}$ & 52.50 & 77.90 & 67.80 & 79.50 & 83.40 & 80.10 & - \\
        MMBERT~\cite{MMBert_model}  &ISBI$_{21}$ & 63.10 & 77.90 & 72.00 & - & - & - & 67.20 \\
        M3AE$^{*}\dag$~\cite{m3ae}  &MICCAI$_{22}$ & 64.80 & 82.72 & 75.61 & 79.22 & 85.10 & 81.53 & 78.40 \\
        M2I2~\cite{M2I2_model}  &ISBI$_{22}$ & 61.80 & 81.60 & 73.70 & 74.70 & \textbf{91.10}& 81.20 & - \\
        ARL$^{*}$~\cite{uar_model}  &MM$_{22}$ & 65.10 & 85.96 & 77.55 & 79.70 & 89.30 & 84.10 & 79.80 \\
        PubMedCLIP~\cite{pubmedclip_model}  &EACL$_{23}$ & 60.10 & 80.00 & 72.10 & 78.40 & 82.50 & 80.10 & - \\
        CPCR~\cite{CPCR}  &TMI$_{23}$ & 60.50 & 80.40 & 72.50 & 80.50 & 84.10 & 81.90 & -\\
        \hline
        LaPA & Ours & \textbf{68.72} & \textbf{86.40} & \textbf{79.38} & \textbf{82.17} & 88.70 & \textbf{84.73} & \textbf{81.60} \\
        \hline \hline
    \end{tabular}
    \caption{The results of the LaPA model and other tested models in VAR-RAD, SLAKE and VQA-2019. $^{*}$ indicates that we tested the results ourselves, which may differ from those reported in the models' original papers. $\dag$ denotes the baseline model. The results for other models were obtained from their original papers. The highest-performing result in each category is highlighted in bold for clarity.}
    \label{table: comparison experiments}
\end{table*}


\begin{table*}[t]
\centering
\begin{tabular}{c|l|ccc|ccc|c}
\hline \hline
\multirow{2}{*}{\#} & \multirow{2}{*}{Method} &\multicolumn{3}{c|}{VQA-RAD} &\multicolumn{3}{c|}{SLAKE}& \multicolumn{1}{c}{VQA-2019}       \\
                    &     & Open  & Closed  & Overall &Open  & Closed & Overall & Overall 
                    \\ \hline
1                   & BL.  & 64.80 & 82.72   & 75.61     &79.22 & 85.10  & 81.53      &78.40      \\
2                   & $\mathbin{+}$GM.$_{w/o\ cs}$ $\And$ LF.  & 68.16 & 84.93   & 78.27 &80.93 & 87.74  & 83.60  &80.80    \\
3                   & $\mathbin{+}$GM.$\And$ LF.   & \textbf{69.27} & 85.29   & 78.94  &81.24 & 87.50  & 83.70  &81.30  \\
4            & $\mathbin{+}$GM.$\And$ LF.$\mathbin{+}$PF.  & 68.72 & \textbf{86.40}   & \textbf{79.38} & \textbf{82.17} & \textbf{88.70}  & \textbf{84.73}   & \textbf{81.60}  \\ \hline
-  & $\Delta$ & {\color{red} $\uparrow$ 3.92} & {\color{red} $\uparrow$ 3.68} & {\color{red} $\uparrow$ 3.77} & {\color{red} $\uparrow$ 2.95} & {\color{red} $\uparrow$ 3.60} & {\color{red} $\uparrow$ 3.20} & {\color{red} $\uparrow$ 3.20} \\ \hline \hline
\end{tabular}
\caption{The ablation study for the LaPA model was conducted on the VQA-RAD, SLAKE, and VQA-2019 datasets to ascertain the contribution of individual components to the overall performance. In this context, GM., LF., and PF. represent the latent prompt generation module, latent prompt fusion module, and prior knowledge fusion module, respectively. The term $w/o\ cs$ denotes the exclusion of the consistency method from the model configuration. The final row delineates the performance enhancement achieved by the LaPA model relative to the established baseline model.}
\label{table: ablation study}
\end{table*}

\subsection{Implementation Details}
\label{subsec: implementation details}

For our model, we adopted the Swin-Transformer~\cite{swin_transformer} as the image extractor model, RoBERTa~\cite{roberta_model} as the language extractor model, the graph attention network~\cite{GAT_model} with eight heads as the GNN model, and utilized six multi-modal fusion blocks. Training was conducted on a single NVIDIA GeForce RTX3090 GPU with 24GB memory, employing half-precision training. Following the approach in M3AE~\cite{m3ae}, we utilized the AdamW optimizer~\cite{adamW} with a learning rate of 5e-6 for optimization. The input images were resized to $384 \times 384$, and the feature dimension was set to 768. Furthermore, we utilized the pre-training weights from the M3AE model, which were pre-trained on the ROCO~\cite{roco} and MedICaT~\cite{medicat} datasets. For evaluation purposes, we report the matching accuracy for both closed-set and open-set questions. The overall metrics are calculated by combining the results from open-set and closed-set questions using coefficients, as outlined in M3AE~\cite{m3ae}.

\subsection{Datasets}
\label{subsec: datasets}

In order to comprehensively evaluate the effectiveness of our proposed method, we conducted experiments on three widely-used Med-VQA benchmarks: VQA-RAD~\cite{rad-vqa}, SLAKE~\cite{slake}, and VQA-2019~\cite{ImageCLEFVQA-Med2019}. The dataset splits provided by existing works, such as M3AE~\cite{m3ae}, were used in our experiments. The questions in VQA-RAD and SLAKE are categorized into two types: open-ended~(free-form) and closed-ended~(YES/NO) forms. VQA-RAD dataset consists of 315 radiology images with 3064 question-answer pairs, and a subset of 451 pairs was used for testing purposes. SLAKE dataset is composed of 642 radiology images, with 14028 question-answer~(QA) pairs. The dataset was divided into a ratio of 70:15:15 for training, validation, and testing, respectively. It it worth noting that we only evaluated the English subset of SLAKE. VQA-2019 dataset comprises 3200 medical images, with 12792 QA pairs for training, 500 images with 2000 QA pairs for validation, and 500 images with 500 QA pairs for testing.

\begin{figure*}[t]
\begin{center}
\includegraphics[width=1\linewidth]{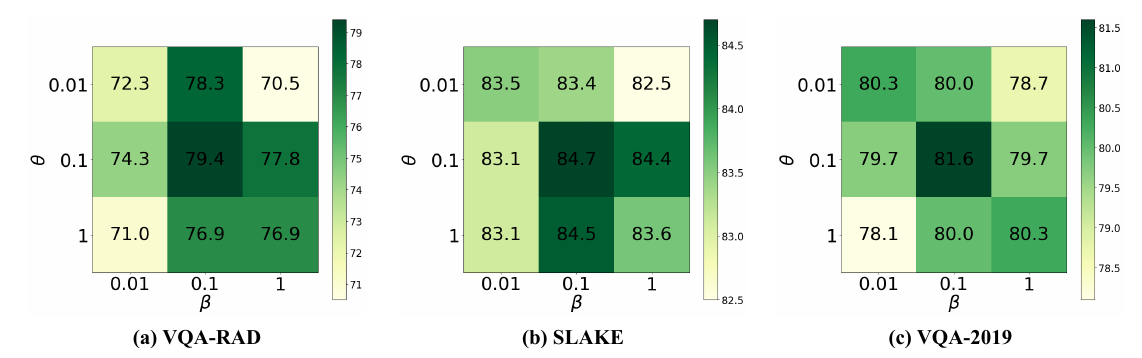}
\end{center}
\vspace{-0.5cm}
\caption{Ablation on the $\theta$ and $\beta$.}
\label{Fig: ablation on the theta and beta}
\end{figure*}


\begin{table*}[t]
    \centering
    \begin{tabular}{ l | c c c | c c c |  c}
        \hline \hline
        \multirow{2}{*}{Interact Order} 
        &\multicolumn{3}{c}{{VQA-RAD}}\vline
        & \multicolumn{3}{c}{{SLAKE}}\vline
        & \multicolumn{1}{c}{{VQA-2019}} \\
         & Open & Closed & Overall & Open & Closed & Overall & Overall \\
        \midrule
         I.$\Rightarrow$L.$\Rightarrow$MM. & 55.31 & 84.56 & 72.95 & 81.40 & 87.74 & 83.88 & 78.93 \\
        L.$\Rightarrow$I.$\Rightarrow$MM. & \textbf{68.72} & \textbf{86.40} & \textbf{79.38} & \textbf{82.17} & \textbf{88.70} & \textbf{84.73} &\textbf{81.60} \\
        \hline \hline
    \end{tabular}
    \caption{The results of the change in the fusion direction by latent prompt in the latent prompt fusion module. The I., L., and MM. are the abbreviations of image, language, and multi-modal.}
    \label{table: ablation on the fusion order}
\end{table*}

\begin{table*}[h!]
    \centering
    \begin{tabular}{c|ccccccc}
    \hline \hline
    Latent Prompt size  & 4    & 8   & 16      & 32      & 64     & 128       & 256   \\ \hline  
    VQA-RAD  & 77.16& 78.27 & 78.49  & \textbf{79.38}  & 76.94  & 76.49     & 75.61 \\
    SLAKE    & 84.17& 84.35 & 83.88  & \textbf{84.73}  & 84.26  & 84.26     & 84.45 \\
    VQA-2019 & 80.00& 80.53 & 79.47  & \textbf{81.60}  & 79.47  & 78.93     & 80.00 \\ \hline \hline
  \end{tabular}
     \caption{Ablation on the latent prompt size.}
  \label{table: analysis coefficient on the LP size}
\end{table*}

\subsection{Comparison Experiments}
\label{subsec:comparison experiments}

Our proposed LaPA model was benchmarked against eight contemporary state-of-the-art (SOTA) Med-VQA methodologies: BAN~\cite{BAN_model}, CPRD~\cite{CPRD_model}, MMBERT~\cite{MMBert_model}, M3AE~\cite{m3ae}, M2I2~\cite{M2I2_model}, ARL~\cite{uar_model}, PubMedCLIP~\cite{pubmedclip_model} and CPCR~\cite{CPCR}. As delineated in Tab.~\ref{table: comparison experiments}, LaPA consistently surpassed the aforementioned models on all three datasets in the majority of evaluative metrics. Notably, for the VQA-RAD dataset, our model demonstrated a considerable enhancement in performance across all question types, achieving an overall accuracy of 79.38\%, an improvement of 1.83 percentage points over the second-best model. In the SLAKE dataset, LaPA achieved an overall accuracy of 84.73\%, outperforming the runner-up by approximately 0.63 percentage points. For VQA-2019, our model registered a significant overall accuracy of 81.6\%, which represents a 1.8 percentage point augmentation compared to the second-best performing model. The M2I2 model exhibited proficiency in answering closed-ended questions but showed limitations with open-ended question types, potentially attributable to disparities in pre-training datasets. The Q2ATransformer~\cite{q2atransformer} and MUMC~\cite{MUMC} models were precluded from our comparison due to the unavailability of their source code, checkpoints, and pre-training datasets, which hindered reproducibility of their results. Moreover, the MeDVInT~\cite{MedVInT} and LLaVA-Med~\cite{llava-med} models possess a parameter count exceeding 7 billion, nearly 17 times that of our LaPA model (0.405B). Despite some superior results from these models, we posit that the comparison would not be equitable due to the vast difference in model size and complexity. Consequently, these models were also excluded from our comparative analysis.

\subsection{Ablation Study}
\label{subsec:ablation study}


In this section, we present an ablation study designed to evaluate the impact of each module within our proposed methodology. The results are summarized in Tab.~\ref{table: ablation study}, encompassing three benchmark datasets. We utilize the following abbreviations: BL. for baseline, GM. for the latent prompt generation module~(detailed in Section~\ref{subsec: Latent Prompt Generation Module}), LF. for the latent prompt fusion module~(described in Section~\ref{subsec: Multi-modal Feature Fusion Module}), and PF. for the prior knowledge fusion module~(elucidated in Section~\ref{subsec: Knowledge-graph-based Latent Prompt Analysis Module}). The notation $w/o\ cs$ specifies configurations that omit the consistency method, which allows for the assessment of its effectiveness. The concluding line quantifies the enhancement our LaPA model offers over the baseline.


Due to the indirect interaction of the latent prompt generation module with image and language modalities, we investigate its influence by conducting an ablation study in conjunction with the GM. and LF. modules. The comparison between conditions \#1 and \#2 in Tab.~\ref{table: ablation study} demonstrates that the integration of the latent prompt markedly enhances the model's capability in addressing Med-VQA tasks. Further, we examine the efficacy of the consistency method; the comparative improvement of condition \#3 over \#2 underscores its utility. The incorporation of the prior knowledge fusion module further augments model performance~(Comparison \#4 and \#3). Ultimately, the amalgamation of all enhancements into the baseline model culminates in a substantial performance leap, as evidenced in condition \#5. The aggregate improvement across all three benchmarks is nearly 3\% relative to the baseline, as detailed in the concluding line of our ablation analysis.

\begin{table}[t]
    \centering
     \begin{tabular}{c|ccccc}
    \hline \hline
    $\eta$  & 0.01  & 0.05   & 0.1   & 0.5      & 1    \\\hline
    VQA-RAD  & 71.84& 72.28 & \textbf{79.38}  & 72.28  & 72.95    \\
    SLAKE    & 83.69& 83.60 & \textbf{84.73}  & 83.22  & 83.60    \\
    VQA-2019 & 80.27& 78.40 & \textbf{81.60}  & 78.40  & 80.80    \\ \hline \hline
  \end{tabular}
  \caption{Ablation on the $\eta$.}
  \vspace{-0.6cm}
  \label{table: analysis coefficient on the eta}
\end{table}

\begin{figure*}[t]
\begin{center}
\includegraphics[width=1\linewidth]{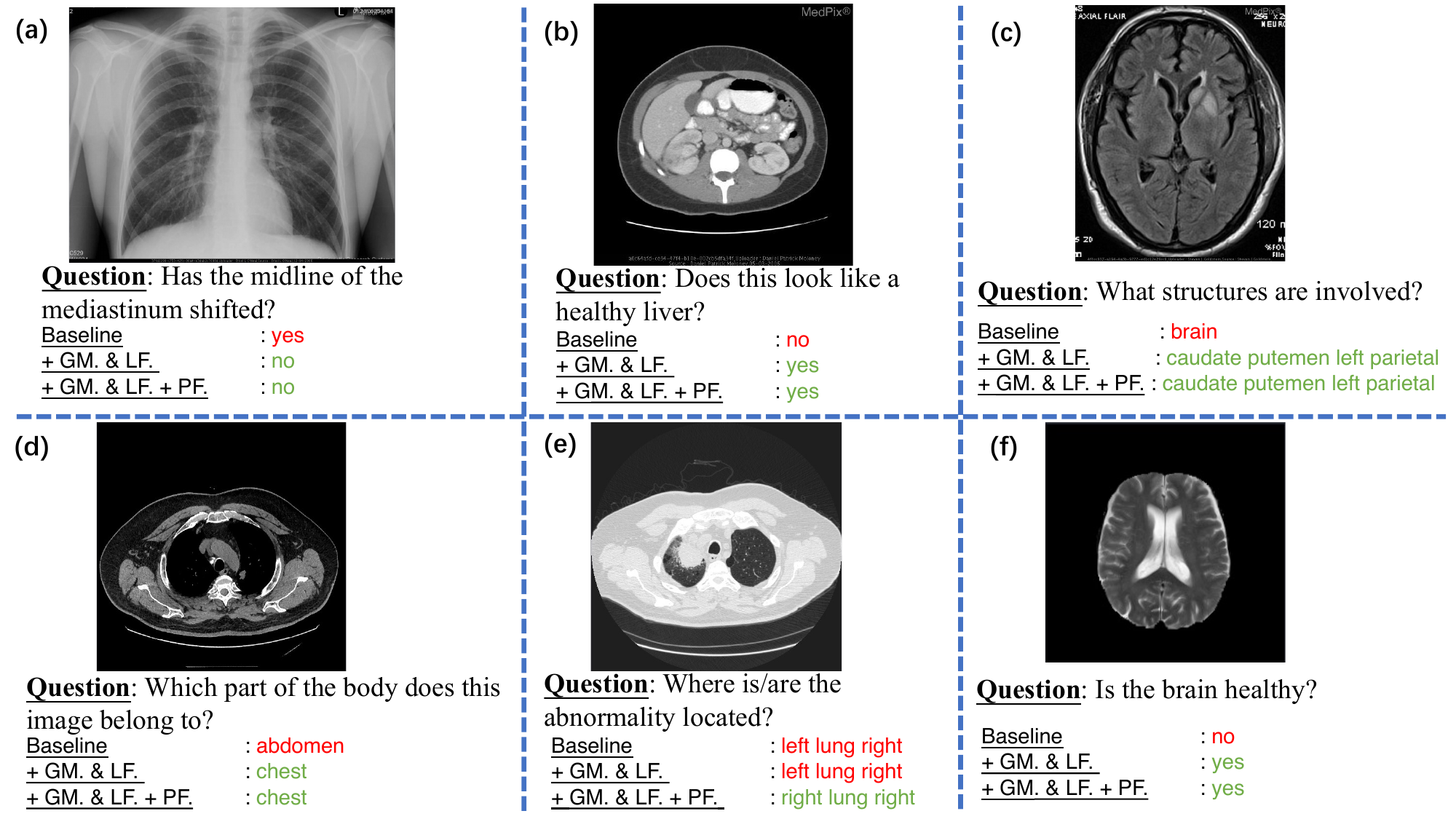}
\end{center}
\caption{Six examples of the LaPA model that use different modules to do the ablation study.  Instances a, b, and c are extracted from the VQA-RAD dataset, whereas instances d, e, and f originate from the SLAKE dataset. Within the provided illustrations, responses are annotated with {\color{green} green} to denote correctness and with {\color{red} red} to signify erroneous predictions by the model. The GM., LF., and PF. are the abbreviations of the latent prompt generation module, latent prompt fusion module, and prior knowledge fusion module.}
\label{Fig: qualitative analysis}
\end{figure*}

\paragraph*{Ablation on the $\theta$ and $\beta$.} 


The hyperparameters $\theta$ and $\beta$ are pivotal in modulating the interaction of cross-modal information, subsequently influencing the accuracy of the final predictive responses. Fig.~\ref{Fig: ablation on the theta and beta} employs a triad of heatmaps to elucidate the effects of various $\theta$ and $\beta$ coefficients on the fusion of cross-modal image and language features within three benchmark datasets. With the coefficient for the latent prompt ($\alpha$) held constant at 1 and the latent prompt size fixed at 32, we systematically vary $\theta$ and $\beta$ from 0.01, through 0.1, to 1 to assess their impact on model performance. The visual representation in Fig.~\ref{Fig: ablation on the theta and beta} indicates that the combination of $\beta = 0.1$ and $\theta = 0.1$ is optimal across all three evaluated datasets.

\paragraph*{Ablation on the interaction order.} 

The sequence of interactions within the latent prompt fusion module exerts a direct influence on the efficacy of information extraction via the latent prompts. Tab.~\ref{table: ablation on the fusion order} delineates the impact of various fusion sequences on the accuracy of the resultant outputs. It is observed that the optimal fusion sequence commences with language, subsequently incorporates image modality, and concludes with a multi-modal fusion, thereby yielding the most favorable outcomes.

\paragraph*{Ablation on the latent prompt size.} 

The size of the latent prompt critically determines the parameter count within the latent prompt framework. Tab.~\ref{table: analysis coefficient on the LP size} presents an analysis of how varying the latent prompt size from 4 to 256 influences performance across the three benchmark datasets. Initially, an increase in latent prompt size correlates with enhanced performance across benchmarks. However, a decline in model accuracy is observed when the latent prompt exceeds a size of 32. The optimal performance, as evidenced by accuracy metrics, is achieved with a latent prompt size of 32 across all evaluated datasets. We hypothesize that excessively large latent prompt dimensions may introduce superfluous and potentially disruptive noise into the information extraction process, thereby detrimentally impacting the precision of the final answer prediction in the Med-VQA context.

\paragraph*{Ablation on the $\eta$.} 

The hyperparameter $\eta$ exerts a direct influence on the weighting of the consistency loss within the aggregate loss function. Tab.~\ref{table: analysis coefficient on the eta} illustrates the impact of varying $\eta$ from 0.01 to 1 on the overall performance across three benchmark datasets. The empirical results indicate that setting $\eta$ to 0.01 yields the most favorable outcomes on all three benchmarks. 


\subsection{Qualitative Analysis}
\label{sec: qualitative analysis}

To further elucidate the efficacy of our Latent Prompt Assist~(LaPA) model, a qualitative analysis was conducted on six Medical visual question answering~(Med-VQA) instances, specifically three from the VQA-RAD dataset (cases a, b, c) and three from the SLAKE dataset (cases d, e, f), as depicted in Fig.~\ref{Fig: qualitative analysis}. Examination of cases a, b, c, d, and f reveals that the incorporation of latent prompts facilitates the model in accurately responding to both closed-ended and open-ended queries across the two benchmarks. However, in case e, the model's integration of solely the latent prompt proved insufficient for distinguishing between two highly similar responses. The addition of the prior Knowledge fusion module~(PF.) was instrumental in rectifying the model's response. These six cases collectively demonstrate that our proposed enhancements substantively bolster the model's performance in resolving both closed-ended and open-ended VQA challenges.

\section{Conclusion}

This study introduces a novel Latent Prompt Assist~(LaPA) model designed to enhance the accuracy of responses in the domain of medical visual question answering~(Med-VQA). It employs the latent prompt to filter different modal information and extract clinic-relevant information to assist in predicting the final answer. Our innovative framework entails a latent prompt generation module that synthesizes latent prompts under the constraint of target answer tokens. These prompts are then integrated with both uni-modal and multi-modal information streams to isolate clinical insights. Further, the model incorporates prior knowledge encapsulated in a knowledge graph, detailing disease-organ relationships, to interact with the latent prompt and refine the final answer prediction. Empirical validation of our approach across three well-established benchmarks demonstrates its superiority in generating accurate answers within the Med-VQA context. Looking forward, we aim to deploy the latent prompt mechanism within a large-scale, highly-parameterized model to fully explore the potential of latent prompts in complex inference tasks.

{\small
\bibliographystyle{ieee_fullname}
\bibliography{egbib}

\begin{thebibliography}{10}\itemsep=-1pt

\bibitem{ImageCLEFVQA-Med2019}
Asma~Ben Abacha, Sadid~A Hasan, Vivek~V Datla, Joey Liu, Dina Demner-Fushman, and Henning M{\"u}ller.
\newblock Vqa-med: Overview of the medical visual question answering task at imageclef 2019.
\newblock {\em CLEF (working notes)}, 2(6), 2019.

\bibitem{m3ae}
Zhihong Chen, Yuhao Du, Jinpeng Hu, Yang Liu, Guanbin Li, Xiang Wan, and Tsung-Hui Chang.
\newblock Multi-modal masked autoencoders for medical vision-and-language pre-training.
\newblock In {\em International Conference on Medical Image Computing and Computer-Assisted Intervention}, pages 679--689. Springer, 2022.

\bibitem{uar_model}
Zhihong Chen, Guanbin Li, and Xiang Wan.
\newblock Align, reason and learn: Enhancing medical vision-and-language pre-training with knowledge.
\newblock In {\em Proceedings of the 30th ACM International Conference on Multimedia}, pages 5152--5161, 2022.

\bibitem{MMQ_model}
Tuong Do, Binh~X Nguyen, Erman Tjiputra, Minh Tran, Quang~D Tran, and Anh Nguyen.
\newblock Multiple meta-model quantifying for medical visual question answering.
\newblock In {\em Medical Image Computing and Computer Assisted Intervention--MICCAI 2021: 24th International Conference, Strasbourg, France, September 27--October 1, 2021, Proceedings, Part V 24}, pages 64--74. Springer, 2021.

\bibitem{pubmedclip_model}
Sedigheh Eslami, Gerard de Melo, and Christoph Meinel.
\newblock Does clip benefit visual question answering in the medical domain as much as it does in the general domain?
\newblock {\em arXiv preprint arXiv:2112.13906}, 2021.

\bibitem{vqamix}
Haifan Gong, Guanqi Chen, Mingzhi Mao, Zhen Li, and Guanbin Li.
\newblock Vqamix: Conditional triplet mixup for medical visual question answering.
\newblock {\em IEEE Transactions on Medical Imaging}, 41(11):3332--3343, 2022.

\bibitem{BCE_loss}
Irving~John Good.
\newblock Rational decisions.
\newblock {\em Journal of the Royal Statistical Society: Series B}, 14(1):107--114, 1952.

\bibitem{COMG_model}
Tiancheng Gu, Dongnan Liu, Zhiyuan Li, and Weidong Cai.
\newblock Complex organ mask guided radiology report generation.
\newblock In {\em Proceedings of the IEEE/CVF Winter Conference on Applications of Computer Vision}, pages 7995--8004, 2024.

\bibitem{ppt_model}
Yuxian Gu, Xu Han, Zhiyuan Liu, and Minlie Huang.
\newblock {PPT}: Pre-trained prompt tuning for few-shot learning.
\newblock In Smaranda Muresan, Preslav Nakov, and Aline Villavicencio, editors, {\em Proceedings of the 60th Annual Meeting of the Association for Computational Linguistics (Volume 1: Long Papers)}, 2022.

\bibitem{MMBert_model}
Yash Khare, Viraj Bagal, Minesh Mathew, Adithi Devi, U~Deva Priyakumar, and CV Jawahar.
\newblock Mmbert: Multimodal bert pretraining for improved medical vqa.
\newblock In {\em 2021 IEEE 18th International Symposium on Biomedical Imaging}, pages 1033--1036. IEEE, 2021.

\bibitem{BAN_model}
Jin-Hwa Kim, Jaehyun Jun, and Byoung-Tak Zhang.
\newblock {Bilinear Attention Networks}.
\newblock In {\em Advances in Neural Information Processing Systems 31}, pages 1571--1581, 2018.

\bibitem{rad-vqa}
Jason~J Lau, Soumya Gayen, Asma Ben~Abacha, and Dina Demner-Fushman.
\newblock A dataset of clinically generated visual questions and answers about radiology images.
\newblock {\em Scientific data}, 5(1):1--10, 2018.

\bibitem{llava-med}
Chunyuan Li, Cliff Wong, Sheng Zhang, Naoto Usuyama, Haotian Liu, Jianwei Yang, Tristan Naumann, Hoifung Poon, and Jianfeng Gao.
\newblock Llava-med: Training a large language-and-vision assistant for biomedicine in one day.
\newblock {\em Advances in Neural Information Processing Systems}, 36, 2024.

\bibitem{MUMC}
Pengfei Li, Gang Liu, Jinlong He, Zixu Zhao, and Shenjun Zhong.
\newblock Masked vision and language pre-training with unimodal and multimodal contrastive losses for medical visual question answering.
\newblock In {\em International Conference on Medical Image Computing and Computer-Assisted Intervention}, pages 374--383. Springer, 2023.

\bibitem{M2I2_model}
Pengfei Li, Gang Liu, Lin Tan, Jinying Liao, and Shenjun Zhong.
\newblock Self-supervised vision-language pretraining for medial visual question answering.
\newblock In {\em 2023 IEEE 20th International Symposium on Biomedical Imaging}, pages 1--5. IEEE, 2023.

\bibitem{lin2023medical}
Zhihong Lin, Donghao Zhang, Qingyi Tao, Danli Shi, Gholamreza Haffari, Qi Wu, Mingguang He, and Zongyuan Ge.
\newblock Medical visual question answering: A survey.
\newblock {\em Artificial Intelligence in Medicine}, page 102611, 2023.

\bibitem{CPRD_model}
Bo Liu, Li-Ming Zhan, and Xiao-Ming Wu.
\newblock Contrastive pre-training and representation distillation for medical visual question answering based on radiology images.
\newblock In {\em Medical Image Computing and Computer Assisted Intervention--MICCAI 2021: 24th International Conference, Strasbourg, France, September 27--October 1, 2021, Proceedings, Part II 24}, pages 210--220. Springer, 2021.

\bibitem{slake}
Bo Liu, Li-Ming Zhan, Li Xu, Lin Ma, Yan Yang, and Xiao-Ming Wu.
\newblock Slake: A semantically-labeled knowledge-enhanced dataset for medical visual question answering.
\newblock In {\em 2021 IEEE 18th International Symposium on Biomedical Imaging}, pages 1650--1654. IEEE, 2021.

\bibitem{CPCR}
Bo Liu, Li-Ming Zhan, Li Xu, and Xiao-Ming Wu.
\newblock Medical visual question answering via conditional reasoning and contrastive learning.
\newblock {\em IEEE transactions on medical imaging}, 42(5):1532--1545, 2022.

\bibitem{roberta_model}
Yinhan Liu, Myle Ott, Naman Goyal, Jingfei Du, Mandar Joshi, Danqi Chen, Omer Levy, Mike Lewis, Luke Zettlemoyer, and Veselin Stoyanov.
\newblock Roberta: A robustly optimized bert pretraining approach.
\newblock {\em arXiv preprint arXiv:1907.11692}, 2019.

\bibitem{q2atransformer}
Yunyi Liu, Zhanyu Wang, Dong Xu, and Luping Zhou.
\newblock Q2atransformer: Improving medical vqa via an answer querying decoder.
\newblock In {\em International Conference on Information Processing in Medical Imaging}, pages 445--456. Springer, 2023.

\bibitem{swin_transformer}
Ze Liu, Yutong Lin, Yue Cao, Han Hu, Yixuan Wei, Zheng Zhang, Stephen Lin, and Baining Guo.
\newblock Swin transformer: Hierarchical vision transformer using shifted windows.
\newblock In {\em Proceedings of the IEEE/CVF International Conference on Computer Vision}, 2021.

\bibitem{adamW}
Ilya Loshchilov and Frank Hutter.
\newblock Decoupled weight decay regularization.
\newblock In {\em ICLR}, 2018.

\bibitem{MEVF_model}
Binh~D Nguyen, Thanh-Toan Do, Binh~X Nguyen, Tuong Do, Erman Tjiputra, and Quang~D Tran.
\newblock Overcoming data limitation in medical visual question answering.
\newblock In {\em Medical Image Computing and Computer Assisted Intervention--MICCAI 2019: 22nd International Conference, Shenzhen, China, October 13--17, 2019, Proceedings, Part IV 22}, pages 522--530. Springer, 2019.

\bibitem{roco}
Obioma Pelka, Sven Koitka, Johannes R{\"u}ckert, Felix Nensa, and Christoph~M Friedrich.
\newblock Radiology objects in context (roco): a multimodal image dataset.
\newblock In {\em Intravascular Imaging and Computer Assisted Stenting and Large-Scale Annotation of Biomedical Data and Expert Label Synthesis: 7th Joint International Workshop, CVII-STENT 2018 and Third International Workshop, LABELS 2018, Held in Conjunction with MICCAI 2018, Granada, Spain, September 16, 2018, Proceedings 3}, pages 180--189. Springer, 2018.

\bibitem{Prompt_for_extraction}
Jiaren Peng, Wenzhong Yang, Fuyuan Wei, and Liang He.
\newblock Prompt for extraction: Multiple templates choice model for event extraction.
\newblock {\em Knowledge-Based Systems}, page 111544, 2024.

\bibitem{prompt_extract}
Morgan~D. Polak, M.P.
\newblock Extracting accurate materials data from research papers with conversational language models and prompt engineering.
\newblock {\em Nat Commun 15, 1569 (2024).}, 2024.

\bibitem{medicat}
Sanjay Subramanian, Lucy~Lu Wang, Sachin Mehta, Ben Bogin, Madeleine van Zuylen, Sravanthi Parasa, Sameer Singh, Matt Gardner, and Hannaneh Hajishirzi.
\newblock Medicat: A dataset of medical images, captions, and textual references.
\newblock {\em arXiv preprint arXiv:2010.06000}, 2020.

\bibitem{rgrg_model}
Tim Tanida, Philip M{\"u}ller, Georgios Kaissis, and Daniel Rueckert.
\newblock Interactive and explainable region-guided radiology report generation.
\newblock In {\em Proceedings of the IEEE/CVF Conference on Computer Vision and Pattern Recognition}, pages 7433--7442, 2023.

\bibitem{transformer_model}
Ashish Vaswani, Noam Shazeer, Niki Parmar, Jakob Uszkoreit, Llion Jones, Aidan~N Gomez, {\L}ukasz Kaiser, and Illia Polosukhin.
\newblock Attention is all you need.
\newblock {\em Advances in neural information processing systems}, 30, 2017.

\bibitem{GAT_model}
Petar Veli{\v{c}}kovi{\'c}, Guillem Cucurull, Arantxa Casanova, Adriana Romero, Pietro Lio, and Yoshua Bengio.
\newblock Graph attention networks.
\newblock {\em arXiv preprint arXiv:1710.10903}, 2017.

\bibitem{SAN_model}
Zichao Yang, Xiaodong He, Jianfeng Gao, Li Deng, and Alex Smola.
\newblock Stacked attention networks for image question answering.
\newblock In {\em Proceedings of the IEEE conference on computer vision and pattern recognition}, pages 21--29, 2016.

\bibitem{zhang2022type}
Anda Zhang, Wei Tao, Ziyan Li, Haofen Wang, and Wenqiang Zhang.
\newblock Type-aware medical visual question answering.
\newblock In {\em ICASSP 2022-2022 IEEE International Conference on Acoustics, Speech and Signal Processing}, pages 4838--4842. IEEE, 2022.

\bibitem{MedVInT}
Xiaoman Zhang, Chaoyi Wu, Ziheng Zhao, Weixiong Lin, Ya Zhang, Yanfeng Wang, and Weidi Xie.
\newblock Pmc-vqa: Visual instruction tuning for medical visual question answering.
\newblock {\em arXiv preprint arXiv:2305.10415}, 2023.

\bibitem{lpt_ref}
Yubo Zhang, Xingxing Zhang, Xun Wang, Si qing Chen, and Furu Wei.
\newblock Latent prompt tuning for text summarization, 2022.

\bibitem{graph_neural_network}
Jie Zhou, Ganqu Cui, Shengding Hu, Zhengyan Zhang, Cheng Yang, Zhiyuan Liu, Lifeng Wang, Changcheng Li, and Maosong Sun.
\newblock Graph neural networks: A review of methods and applications.
\newblock {\em AI open}, 1:57--81, 2020.

\bibitem{conditional_prompt_learning}
Kaiyang Zhou, Jingkang Yang, Chen~Change Loy, and Ziwei Liu.
\newblock Conditional prompt learning for vision-language models.
\newblock In {\em Proceedings of the IEEE/CVF conference on computer vision and pattern recognition}, pages 16816--16825, 2022.

\bibitem{zhou2022learning}
Kaiyang Zhou, Jingkang Yang, Chen~Change Loy, and Ziwei Liu.
\newblock Learning to prompt for vision-language models.
\newblock {\em International Journal of Computer Vision}, 130(9):2337--2348, 2022.

\end{thebibliography}
}

\end{document}